\documentclass{article}
\usepackage{iclr2026_conference,times}

\usepackage{multirow}
\usepackage[utf8]{inputenc}
\usepackage[T1]{fontenc}
\usepackage{hyperref}
\usepackage{url}
\usepackage{booktabs}
\usepackage{amsfonts}
\usepackage{nicefrac}
\usepackage{microtype}
\usepackage{xcolor}
\usepackage{amsmath}
\usepackage{subcaption}
\usepackage{graphicx}
\usepackage{pifont}
\title{Copy Less, Ground More: Overcoming Repetitive Copying in Long-Context Reasoning via Evidence-Aware Reinforcement Learning}

\author{Lizhe Fang${{}^{1,2}}$\thanks{Part of work done during internship at Qwen Team of Alibaba Group.}  \quad Weizhou Shen${{}^2}$ \quad Tianyi Tang${{}^2}$ \quad Yisen Wang${{}^{1}}$\thanks{Corresponding Author: Yisen Wang (yisen.wang@pku.edu.cn).}  \\
${{}^1}$ Peking University\\
${{}^2}$ Alibaba Group ATH\,
\raisebox{-0.2\height}{\includegraphics[height=1.2em]{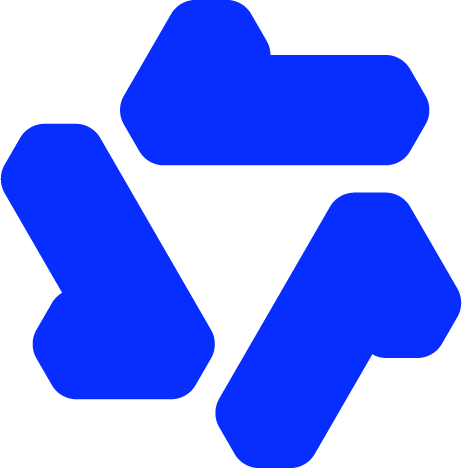}} \\
}

\iclrfinalcopy
\begin{document}

\maketitle
\lhead{arXiv Preprint}

\begin{abstract}
Large language models that generate step-by-step reasoning traces have achieved strong performance on complex tasks, and extending them to long-context settings has emerged as an important frontier. However, we identify a critical failure mode in this regime: \emph{repetitive copying}, where models extensively copy text from the input into their reasoning traces rather than productively solving the problem. We show that this behavior is pervasive across frontier long-context LLMs and intensifies with context length. By separating each prompt into task-relevant key evidence and irrelevant distractor context, we further show that the root cause is insufficient grounding: models copy from the prompt indiscriminately, and those that fail to focus on key evidence are far more likely to answer incorrectly. Motivated by this diagnosis, we propose GEAR (Grounding Evidence-Aware Reward), a reward shaping method that augments the accuracy signal with a grounding reward for overlap with key evidence and a distractor penalty for overlap with irrelevant context. To enable GEAR on natural-language data, we develop an automated pipeline that constructs evidence-annotated training data from arbitrary documents. We validate GEAR across multiple model scales and benchmarks, showing consistent improvements of up to +4.6 average points over standard RL with accuracy-based rewards, with larger gains at longer contexts, while also reducing repetitive copying and thinking length. Our findings suggest that, even as long-context evaluation shifts from simple retrieval toward complex reasoning, accurate grounding in relevant evidence remains an indispensable capability with substantial room for improvement.
\end{abstract}

\section{Introduction}
\label{sec:intro}

Recent large language models (LLMs) equipped with explicit thinking capabilities have demonstrated remarkable performance on complex reasoning tasks. By generating step-by-step reasoning traces before arriving at a final answer, thinking models such as the Qwen3 series \citep{yang2025qwen3}, the Claude series, and the DeepSeek series \citep{guo2025deepseek} consistently outperform their non-thinking counterparts. Meanwhile, the evaluation of long-context LLMs has increasingly shifted from simple retrieval tasks toward complex reasoning over extended inputs \citep{bai2025longbench, hsieh2024ruler, chen2026longrlvr}, making these thinking models a natural fit for the frontier of long-context reasoning.

Despite this progress, we identify a critical failure pattern when thinking models are applied to \emph{long-context reasoning}: \textbf{repetitive copying}. When reasoning over long inputs, models frequently copy large passages directly from the prompt into their thinking traces. This behavior is pervasive, appearing across seven frontier long-context LLMs that include both proprietary APIs and open-weight checkpoints, and it becomes more severe as context length increases. Crucially, repetitive copying not only reduces accuracy but also substantially increases reasoning length, as the model spends its token budget reproducing prompt content rather than solving the task.

To understand the root cause of this behavior, we conduct a systematic analysis and find that the core issue is \textbf{insufficient grounding}: models copy prompt content indiscriminately because they fail to distinguish task-relevant evidence from irrelevant distractor context. We quantify how selectively a model copies from task-relevant vs.\ irrelevant content and show that correctly answered samples exhibit significantly more selective copying than incorrect ones. These findings suggest that copying from the prompt is not inherently harmful. The problem instead lies in indiscriminate copying: models that focus on key evidence are more likely to succeed, whereas those that copy broadly from the context are more likely to fail.

Motivated by this diagnosis, we propose \textbf{GEAR} (\textbf{G}rounding \textbf{E}vidence-\textbf{A}ware \textbf{R}eward), a reward shaping method that augments the standard accuracy signal with two complementary components. A \emph{grounding reward} encourages the model to engage with key evidence by measuring $n$-gram overlap between the reasoning trace and annotated support spans. A \emph{distractor penalty} discourages indiscriminate copying by penalizing overlap with irrelevant context. To extend GEAR beyond synthetic benchmarks where evidence annotations are readily available, we further develop an automated three-stage pipeline that constructs evidence-annotated QA pairs from arbitrary long documents.

We train three Qwen3.5 models across different scales with GSPO \citep{zheng2025group} using the GEAR reward on a mixture of 3.2k long-context samples from both synthetic tasks and natural-language documents. On five held-out benchmarks, GEAR consistently improves over standard RL with accuracy-based rewards by up to +4.6 average points, while simultaneously reducing repetitive copying and thinking length.

Our contributions are as follows:
\begin{itemize}
    \item We identify the \emph{repetitive copying} problem in long-context LLM reasoning, showing that it is prevalent across models and negatively correlated with accuracy. We further trace its root cause to insufficient evidence grounding, as models copy indiscriminately instead of selectively engaging with task-relevant information.
    \item We propose GEAR, a simple and effective reward shaping method that combines a grounding reward with a distractor penalty to encourage selective evidence engagement during RL training.
    \item We develop an automated data construction pipeline that generates evidence-annotated long-context QA pairs from arbitrary document corpora, enabling GEAR to scale beyond synthetic benchmarks to natural-language data.
    \item We validate GEAR on five diverse benchmarks, showing consistent improvements over standard RL with accuracy-based rewards and confirming that the gains stem from reduced indiscriminate copying and improved grounding.
\end{itemize}

\section{Related Work}

\textbf{Reinforcement learning for reasoning.}
The success of DeepSeek-R1 \citep{guo2025deepseek} demonstrated that strong reasoning capabilities can emerge from RL training with verifiable rewards alone, without supervised fine-tuning. This reinforcement learning with verifiable rewards (RLVR) paradigm has since been refined along several axes. GRPO \citep{shao2024deepseekmath} replaces the critic network with group-relative advantage estimation, reducing memory overhead. DAPO \citep{yu2026dapo} introduces asymmetric clipping and dynamic sampling for more stable long-CoT training. GSPO \citep{zheng2025group} shifts from token-level to sequence-level importance ratios, addressing variance issues in MoE models. On the data side, s1 \citep{muennighoff2025s1} shows that as few as 1,000 curated traces can activate latent reasoning ability, while RLPR \citep{yu2025rlpr} uses the model's own generation probability as a graded reward to extend RLVR beyond math domains. These works focus primarily on short-context reasoning; our work addresses the distinct challenges that arise when RL is applied to long-context tasks.

\textbf{Long-context RL training.}
Several concurrent works extend RLVR to long-context settings. LongRLVR \citep{chen2026longrlvr} proves that outcome-only rewards cause vanishing gradients for contextual grounding and augments the answer reward with a verifiable context reward based on chunk-level F-scores. GoLongRL \citep{lv2026golongrl} takes a capability-oriented approach, constructing a 23K-sample dataset spanning 9 task types with heterogeneous reward functions and proposing task-level reward normalization. QwenLong \citep{wan2025qwenlong,shen2025qwenlong} combines curriculum-guided RL with difficulty-aware sampling, while LongR \citep{ping2026longr} introduces a contextual density reward that measures information gain from cited context. LoongRL \citep{wang2025loongrl} pads short-context multi-hop QA with distractors to create long-context training data. Our work differs from these approaches in both diagnosis and method: we identify \emph{repetitive copying} as a specific failure mode of long-context reasoning and design a reward that directly targets this behavior by distinguishing key evidence from distractor context at the $n$-gram level, rather than operating on coarse chunk identifiers.

\textbf{Repetition and overthinking in LLMs.}
Repetitive degeneration in neural text generation has been studied from multiple angles. \citet{li2023repetition} trace degeneration to repetitive training data, while \citet{yao2025understanding} identify specific ``repetition features'' in model activations. \citet{mahaut2025repetitions} show that repetition arises from distinct mechanisms including uncertainty and overconfidence. In the context of reasoning models, the ``overthinking'' problem, in which models generate excessively long reasoning traces, has received growing attention \citep{chen2024not, dai2026s}, with \citet{wu2025more} demonstrating an inverted-U relationship between reasoning length and accuracy. Our work connects these threads by showing that in long-context settings, the dominant form of excessive reasoning is not aimless elaboration but \emph{direct copying from the prompt}, and that this behavior is directly linked to insufficient evidence grounding. Separately, \citet{fang2025wrong} show that not all tokens contribute equally to long-context understanding, suggesting that the impact of repetitive generation may depend on \emph{what} content is being repeated rather than repetition alone.

\section{The Repetitive Copying Issue in Long-Context Reasoning}

In this section, we present a systematic empirical study of repetitive copying in thinking LLMs. Our analysis proceeds in three stages. First, we quantify how prevalent this behavior is across models and context lengths (Section~\ref{sec:prevalence}). Second, we establish that higher copying rates correlate with lower task accuracy and inflated reasoning length (Section~\ref{sec:harmful}). Third, we investigate the root cause by examining \emph{what} the model copies (Section~\ref{sec:grounding}).

\subsection{Experimental Setup}

\textbf{Metrics.} We quantify repetitive copying using \textit{overlap rate}, which measures how much of a model's output is copied directly from the input prompt. We define $y = (y_1, \ldots, y_m)$ as the reasoning trace (model output) and collect all distinct $n$-grams from the prompt $x$ into a lookup set $\mathcal{N}_n(x)$. We then scan through $y$ with a sliding window of size $n$: for each position $i$, the window $(y_i, \ldots, y_{i+n-1})$ is marked as a match if it appears in $\mathcal{N}_n(x)$. The overlap rate is the match rate across all positions:

\begin{equation}
\label{eq:overlap}
\text{Overlap}_n(y \| x) = \frac{1}{m - n + 1}\sum_{i=1}^{m-n+1}
  \mathbf{1}\!\left[(y_i, \ldots, y_{i+n-1}) \in \mathcal{N}_n(x)\right].
\end{equation}

We report results for both $n=3$ and $n=10$: 3-grams capture overall copying volume, while 10-grams isolate longer contiguous spans that unambiguously indicate deliberate transcription.

\textbf{Models.} We evaluate seven thinking LLMs spanning proprietary APIs and open-weight checkpoints: Claude-Opus-4.5, DeepSeek-V3.2 \citep{liu2025deepseek}, Qwen-3.5-Plus \citep{yang2025qwen3}, GLM-4.7 \citep{glm2024chatglm}, QwQ-32B \citep{qwq32b}, Qwen3.5-35B-A3B, and Qwen3.5-9B. All models produce explicit reasoning traces before generating a final answer, allowing direct inspection of their intermediate reasoning for copying behavior.

\textbf{Task.} We study repetitive copying on GSM-Infinite \citep{zhou2025gsm}, a procedurally generated long-context arithmetic reasoning benchmark that extends GSM8K by embedding word problems within a large body of numerical descriptions. Each problem requires 10--20 sequential operations, and the model must identify the relevant numerical conditions scattered across the context and chain them correctly. Because problems and filler content are generated programmatically, both the character positions of the relevant conditions (support indices) and the ground-truth reasoning chains are known exactly, enabling precise measurement of \emph{what} the model copies, i.e., whether it targets task-relevant evidence or merely reproduces filler text. We construct datasets at context scales of 8k, 16k, 32k, and 64k tokens.

\subsection{Repetitive Copying is Prevalent among LLMs}
\label{sec:prevalence}

We begin by measuring how much reasoning models copy from the input and how this behavior changes with context length.

\begin{figure}[t]
\centering
\begin{subfigure}[b]{0.49\textwidth}
    \includegraphics[width=\textwidth]{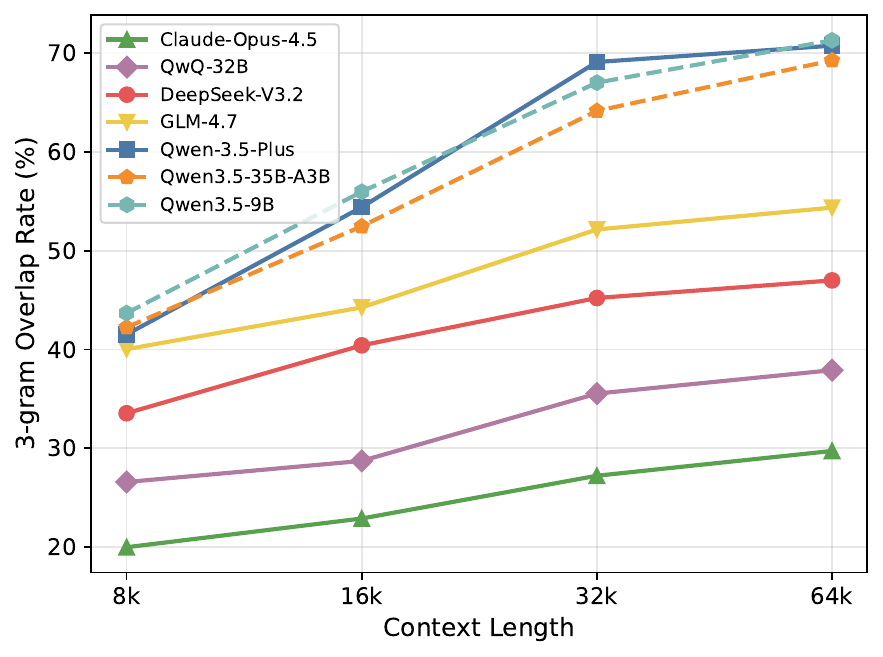}
    \caption{$n = 3$}
    \label{fig:prev_3gram}
\end{subfigure}
\hfill
\begin{subfigure}[b]{0.49\textwidth}
    \includegraphics[width=\textwidth]{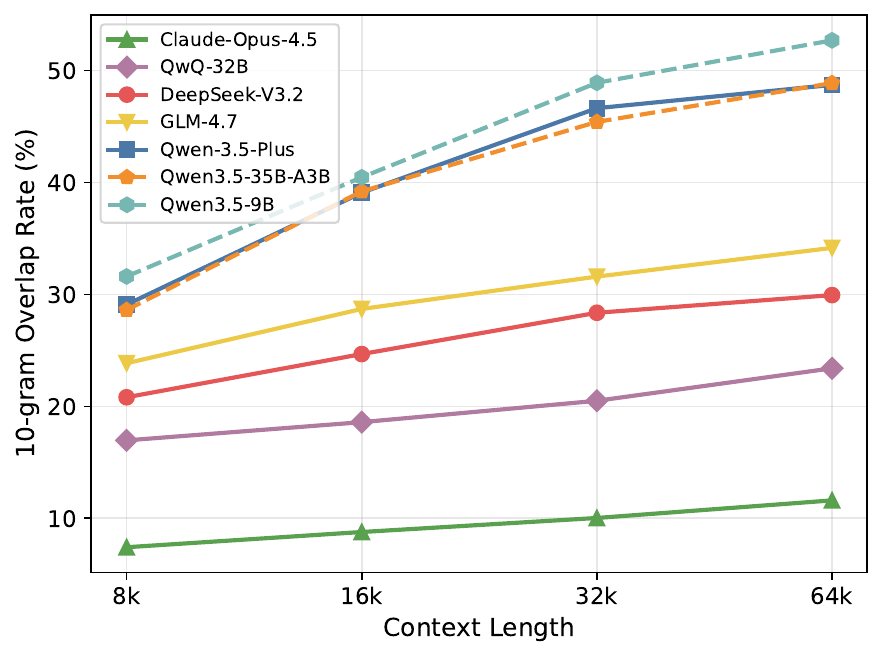}
    \caption{$n = 10$}
    \label{fig:prev_10gram}
\end{subfigure}
\caption{$n$-gram overlap rate on GSM-Infinite as a function of context length across seven models. All models copy more as context grows, and the high 10-gram overlap confirms that copied spans are long contiguous passages.}
\label{fig:prevalence}
\end{figure}

Figure~\ref{fig:prevalence} reports overlap rates on GSM-Infinite across four context scales (8k--64k). All models exhibit substantial copying even at the shortest context. At 8k, 3-gram overlap rates range from 20.0\% (Claude) to 43.7\% (Qwen3.5-9B), indicating that a significant portion of reasoning traces consists of copied prompt content. The problem worsens uniformly as context grows: Qwen-3.5-Plus increases from 41.5\% to 70.8\% at 64k, and even the least affected model (Claude) rises from 20.0\% to 29.7\%. The trend is consistent across all seven models, suggesting that longer inputs systematically elicit more copying during reasoning.

Comparing the 3-gram and 10-gram panels reveals that a significant fraction of copied content consists of long contiguous spans. The 10-gram overlap of Qwen3.5-9B reaches 52.7\% at 64k, meaning that over half of all 10-gram windows in the reasoning trace are direct copies from the prompt, providing strong evidence of deliberate transcription rather than incidental lexical overlap.

\subsection{Repetitive Copying Harms Task Performance}
\label{sec:harmful}

Having established that repetitive copying is widespread across all models (Section~\ref{sec:prevalence}), we now examine its relationship with task performance. We use DeepSeek-V3.2 on GSM-Infinite at 64k context as a representative case study; results for additional models are provided in Appendix~\ref{app:overlap_models}.

\begin{figure}[t]
\centering
\begin{subfigure}[b]{0.55\textwidth}
    \includegraphics[width=\textwidth]{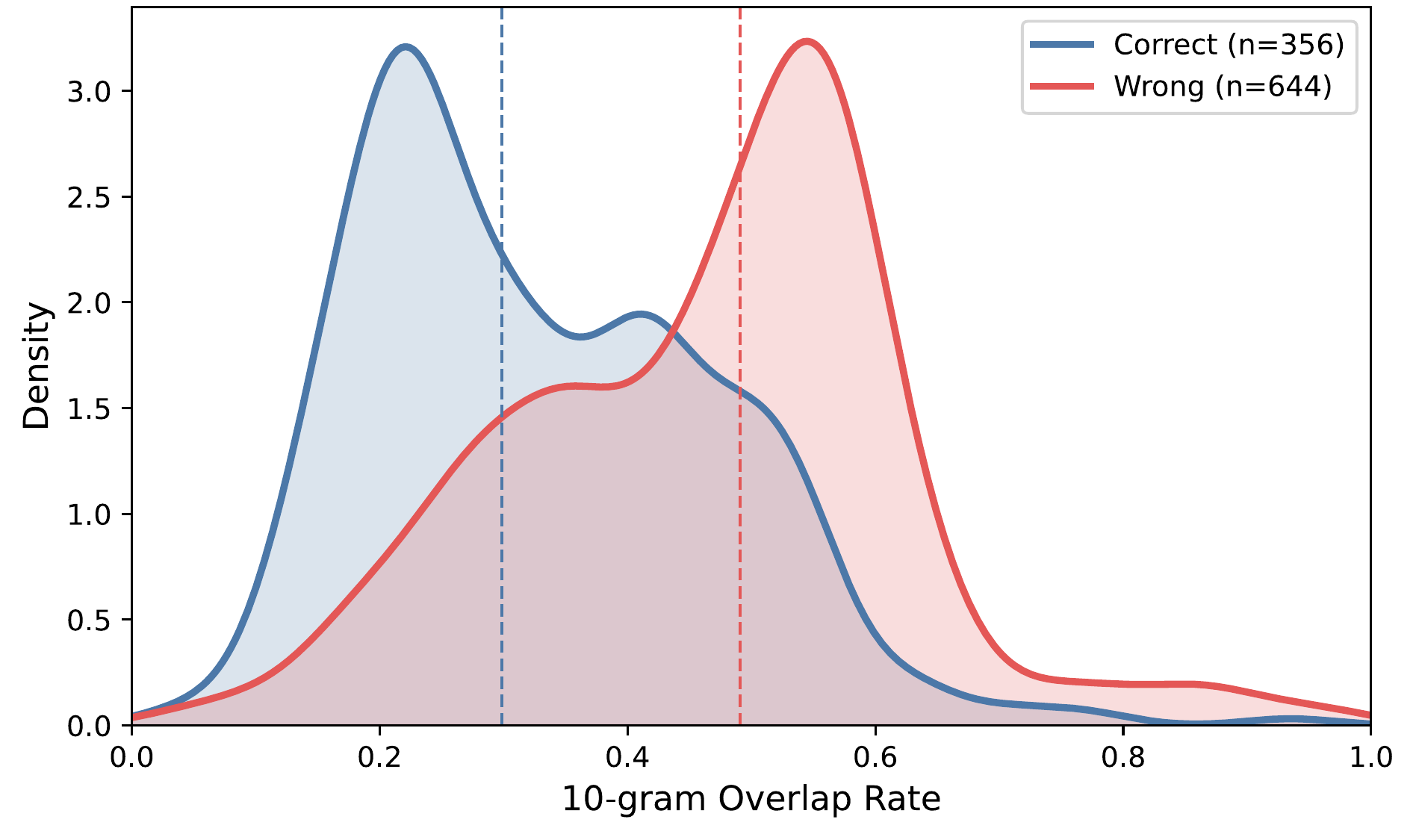}
    \caption{Overlap rate distributions}
    \label{fig:overlap_kde}
\end{subfigure}
\hfill
\begin{subfigure}[b]{0.44\textwidth}
    \includegraphics[width=\textwidth]{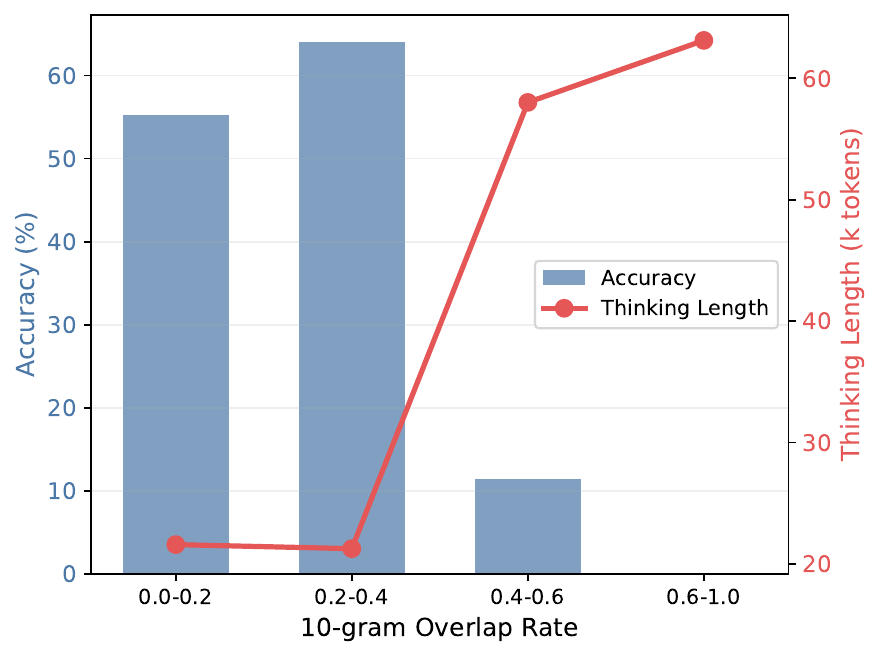}
    \caption{Accuracy and thinking length vs.\ overlap}
    \label{fig:overlap_acc_len}
\end{subfigure}
\caption{Repetitive copying harms both accuracy and reasoning efficiency (DeepSeek-V3.2 on GSM-Infinite, 64k context, 10-gram overlap). (a) Correctly answered samples exhibit markedly lower overlap rates. (b) As overlap rate increases, accuracy drops and thinking length (in tokens) grows sharply.}
\label{fig:harmful}
\end{figure}

\textbf{Higher overlap correlates with lower accuracy and longer reasoning.} We compare the 10-gram overlap distributions of correctly and incorrectly answered samples (Figure~\ref{fig:overlap_kde}). The two groups are clearly separated: correctly answered samples exhibit far less copying from the prompt than incorrect ones. This suggests that excessive copying is not a productive strategy: rather than helping the model locate relevant information, it floods the reasoning trace with unreprocessed prompt content.

The damage becomes even clearer when we examine the joint effect on accuracy and reasoning length (Figure~\ref{fig:overlap_acc_len}). Below an overlap rate of 0.4, accuracy remains at 55--64\% and thinking length stays around 21k tokens. Beyond this threshold, accuracy collapses to 11\% and eventually 0\%, while thinking length nearly triples to over 58k tokens. In other words, the model spends its token budget transcribing the input rather than reasoning about it, and produces longer but less useful traces as a result. Importantly, this relationship is not merely a confound of problem difficulty: the same pattern holds when we group samples by the number of required reasoning steps (Appendix~\ref{app:difficulty}).

\subsection{Repetitive Copying Reflects Insufficient Grounding}
\label{sec:grounding}

Section~\ref{sec:harmful} establishes that repetitive copying correlates with lower accuracy, but it does not explain \emph{why}, since copying from the prompt is not inherently harmful. If the model selectively retrieves task-relevant information, it could even be beneficial. The critical question, then, is not \emph{how much} the model copies, but \emph{what} it copies: does it focus on key evidence, or does it reproduce prompt content indiscriminately?

\textbf{Measuring grounding with the key/distractor overlap ratio.} To answer this question, we leverage the procedural construction of GSM-Infinite, where ground-truth support indices identify which portions of the prompt contain the key evidence needed to solve each problem. We split each prompt into key evidence (the annotated support spans) and distractor context (the remainder), and compute the overlap against each part separately. We use $n = 10$ for this analysis, as longer $n$-grams more reliably indicate deliberate transcription rather than incidental lexical overlap (Section~\ref{sec:prevalence}). The \emph{grounding ratio} is defined as:
\begin{equation}
\label{eq:grounding_ratio}
r = \frac{\text{Overlap}_{10}(y \| x^{\mathrm{key}})}{\text{Overlap}_{10}(y \| x^{\mathrm{dist}})}
\end{equation}
A higher grounding ratio indicates that the model's copying is more focused on task-relevant evidence.

\begin{figure}[t]
\centering
\begin{subfigure}[b]{0.49\textwidth}
    \includegraphics[width=\textwidth]{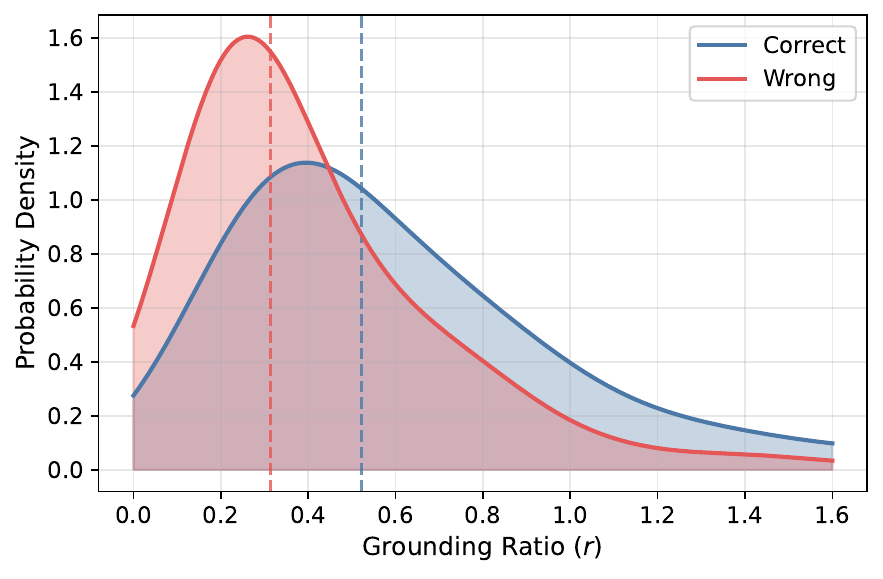}
    \caption{Grounding ratio}
    \label{fig:ratio_kde_a}
\end{subfigure}
\hfill
\begin{subfigure}[b]{0.50\textwidth}
    \includegraphics[width=\textwidth]{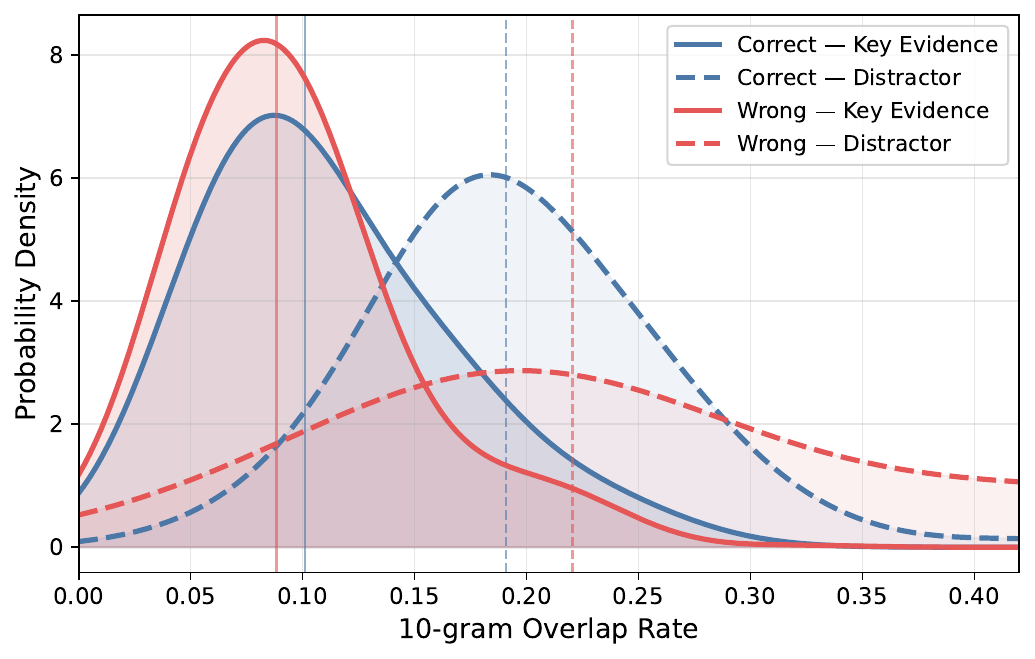}
    \caption{Key evidence vs.\ distractor overlap}
    \label{fig:ratio_kde_b}
\end{subfigure}
\caption{Copying behavior of correctly vs.\ incorrectly answered samples (DeepSeek-V3.2 on GSM-Infinite, 64k context). (a) Probability density of grounding ratio $r$ (Eq.~\ref{eq:grounding_ratio}). Correctly answered samples exhibit higher grounding ratios, indicating more selective copying. (b) 10-gram overlap with key evidence (solid) and distractor context (dashed). Correctly answered samples show higher key evidence overlap and lower distractor overlap, confirming more selective copying from task-relevant evidence. Vertical lines indicate medians.}
\label{fig:ratio_kde}
\end{figure}

Figure~\ref{fig:ratio_kde_a} shows the distribution of grounding ratios for correctly vs.\ incorrectly answered samples. Correctly answered samples exhibit markedly higher grounding ratios, indicating that successful reasoning involves more selective copying from task-relevant evidence. Figure~\ref{fig:ratio_kde_b} decomposes this further: correctly answered samples show both higher key evidence overlap and lower distractor overlap, confirming that accurate reasoning involves selectively engaging with relevant evidence while avoiding irrelevant context. We observe this pattern across five of the six additional models tested (Appendix~\ref{app:grounding}).

These results indicate that the model's long-context reasoning suffers not because it copies from the prompt \emph{per se}, but because it fails to ground its copying in the relevant evidence. In summary, our analysis identifies a prevalent failure mode in long-context reasoning: models extensively copy irrelevant passages from the input into their reasoning traces, crowding out productive problem-solving.


\section{GEAR: Grounding Evidence-Aware Reward}
\label{sec:method}

Our analysis identifies insufficient grounding as the root cause of harmful repetitive copying (Section~\ref{sec:grounding}). We propose \textbf{GEAR} (\textbf{G}rounding \textbf{E}vidence-\textbf{A}ware \textbf{R}eward), a reward shaping method that augments the standard accuracy signal with a grounding reward and a distractor penalty to directly encourage selective evidence engagement during RL training. We first describe the reward design (Section~\ref{sec:reward_design}), then present an automated data construction pipeline that produces the evidence annotations GEAR requires (Section~\ref{sec:data_construction}), and finally detail the training setup (Section~\ref{sec:training_setup}).

\subsection{Reward Design}
\label{sec:reward_design}

Since correctly answered samples exhibit systematically higher overlap with key evidence and lower overlap with distractor context (Section~\ref{sec:grounding}), we design two complementary reward signals that directly target this distinction. Given a prompt $x$ with ground-truth support indices that partition it into key evidence $x^{\mathrm{key}}$ and distractor context $x^{\mathrm{dist}}$, and a model-generated reasoning trace $y$, the GEAR reward is:
\begin{equation}
\label{eq:reward}
R(x, y) = R_{\mathrm{acc}} + \alpha \cdot \text{Overlap}_n(y \| x^{\mathrm{key}}) - \beta \cdot \text{Overlap}_n(y \| x^{\mathrm{dist}})
\end{equation}
where $R_{\mathrm{acc}} \in \{0, 1\}$ is the binary accuracy reward, $\text{Overlap}_n(\cdot)$ is the $n$-gram overlap rate defined in Eq.~\ref{eq:overlap}, and $\alpha$, $\beta$ are non-negative coefficients. In practice, $n$-grams that appear in both $x^{\mathrm{key}}$ and $x^{\mathrm{dist}}$ are excluded from the distractor term to avoid penalizing legitimate references to key evidence.

The grounding reward $R_{\mathrm{ground}} = \alpha \cdot \text{Overlap}_n(y \| x^{\mathrm{key}})$ encourages the model to selectively engage with task-relevant evidence, while the distractor penalty $R_{\mathrm{dist}} = \beta \cdot \text{Overlap}_n(y \| x^{\mathrm{dist}})$ discourages indiscriminate copying of irrelevant context. The two forces are complementary: the grounding term alone cannot prevent the model from copying \emph{everything} (which would also yield high key overlap), and the distractor penalty alone cannot encourage the model to engage with key evidence at all. Notably, this reward requires no external verifier or retrieval module, as it operates entirely on $n$-gram statistics computable from the existing support annotations.

\subsection{Evidence-Annotated Data Construction}
\label{sec:data_construction}

While the support annotations required by GEAR are straightforward for synthetic benchmarks, extending to natural-language data requires an automated pipeline. We describe two complementary data sources below.

\textbf{Structured synthetic data.} We select two procedurally generated benchmarks that require distinct reasoning skills over long contexts. GSM-Infinite \citep{zhou2025gsm} extends GSM8K into a long-context setting by embedding arithmetic word problems within large bodies of numerical descriptions; the model must locate relevant numerical conditions scattered across the context and chain 10--20 sequential operations. PhantomWiki \citep{gong2025phantomwiki} constructs fictional encyclopedias with interlinked entity articles, requiring multi-hop entity tracking to answer questions whose evidence spans multiple articles. We choose these two tasks to cover complementary reasoning types (numerical chaining and entity tracking) so that GEAR learns general evidence-grounding behavior rather than task-specific shortcuts. Because both benchmarks are generated programmatically, their generators natively produce support indices as part of the construction logic, requiring no additional annotation effort.

\textbf{Natural-language data: a three-stage synthesis pipeline.} To extend GEAR beyond synthetic benchmarks, we develop an automated pipeline that generates evidence-annotated QA pairs from arbitrary long documents. The key idea is to reverse the usual annotation workflow: rather than generating a question and then labeling which parts of the document support it, we first sample a small subset of document chunks as the designated evidence region, then constrain question generation to this region alone. This way, support annotations are obtained by construction. Given a source document $d$, the pipeline proceeds as follows:

\begin{enumerate}
    \item \textbf{Document analysis and filtering.} A language model analyzes $d$ to identify its informational characteristics (factual claims, numerical data, causal relationships, comparisons) and determines which question types are appropriate. Documents unsuitable for QA (e.g., structured data, product listings) are filtered out, and the identified question types guide the subsequent generation step.

    \item \textbf{Evidence-constrained QA generation.} The document is segmented into text chunks (1,024 characters each). A random subset of 1--3 chunks is sampled as the \emph{constraint context}, defining the key evidence region; the remainder constitutes the distractor context. A language model then generates a question answerable from the constraint context, along with 1--3 exact evidence quotations and a concise answer.

    \item \textbf{Answer verification.} The generated question is independently re-answered using \emph{only} the constraint context. We retain only pairs where this re-derived answer matches the reference answer, ensuring that the question is unambiguously answerable from the designated evidence.
\end{enumerate}

We restrict answers to single-valued responses (a number or entity phrase, excluding yes/no questions) to support automated reward computation.

\textbf{Training set composition.} Using these two pipelines, we construct a training set of 3,200 problems: 1,000 each from GSM-Infinite and PhantomWiki (structured synthetic), plus 1,200 QA pairs generated from RedPajama-v2 \citep{weber2024redpajama} documents using the natural-language pipeline. The inclusion of both data types is intended to verify that GEAR generalizes across data sources rather than being limited to synthetic benchmarks.

\subsection{Training Setup}
\label{sec:training_setup}

\textbf{Base model and algorithm.} We train three base models: Qwen3.5-9B, Qwen3.5-27B, and Qwen3.5-35B-A3B (a Mixture-of-Experts variant) using Group Sequence Policy Optimization \citep{zheng2025group} with the GEAR reward defined in Eq.~\ref{eq:reward}. We set $n = 3$ for $n$-gram computation, rather than the $n = 10$ used in the diagnostic analysis (Section~\ref{sec:grounding}), as shorter $n$-grams provide denser reward signal that is more effective for RL optimization, and our ablation study confirms this is the optimal setting. We set $\alpha = 0.1$ and $\beta = 0.3$, penalizing distractor copying more aggressively than rewarding key evidence overlap; our sensitivity analysis validates this choice.

\textbf{Training details.} We train for 1 epoch (${\sim}100$ optimizer steps) with a learning rate of $2 \times 10^{-6}$, a batch size of 32, and a rollout group size of 8. The maximum prompt length is 32k tokens and the maximum generation length is 16k tokens. Training data context lengths are uniformly distributed between 16k and 32k tokens.

\section{Experiments}
\label{sec:experiments}

\subsection{Main Results}
\label{sec:main_results}

\textbf{Setup.} We evaluate on five long-context benchmarks that are disjoint from the GEAR training data: Ruler \citep{hsieh2024ruler}, a synthetic retrieval benchmark testing multi-key and multi-value extraction across varying context lengths; LongBench-v2 \citep{bai2025longbench}, a comprehensive benchmark covering diverse long-context understanding tasks; BrowseComp-LC \citep{wei2025browsecomp}, a web browsing comprehension benchmark requiring synthesis of information across long documents; GraphWalks \citep{openai2025gpt41}, a graph-structured reasoning benchmark that requires models to follow multi-hop paths and preserve intermediate state over long textual graph descriptions; and AA-LCR \citep{aalcr2025}, a long-context reasoning benchmark. All benchmarks natively use 128k-token contexts. We report results at two scales: a \emph{32k} subset containing only tasks whose context fits within 32k tokens, and the full \emph{128k} set. Because all AA-LCR instances exceed 32k tokens, this benchmark appears only in the 128k column. For each base model, we compare four configurations: (1) the pretrained Qwen3.5 base model, (2) GSPO trained with accuracy-only reward, (3) GSPO with the grounding reward added ($+ R_{\mathrm{ground}}$), and (4) GSPO with the full GEAR reward ($+ R_{\mathrm{ground}} + R_{\mathrm{dist}}$).

\textbf{GEAR yields consistent gains across benchmarks and model scales.} Table~\ref{tab:main_results} presents results across three model scales and two context lengths. GSPO + GEAR achieves the highest average score in all six model--context combinations. At 32k context, GEAR improves over accuracy-only GSPO by +2.8 (9B), +2.1 (35B-A3B), and +1.5 (27B) points on average. At 128k, the gains are larger: +4.6, +2.8, and +2.1, respectively. This suggests that evidence-aware reward shaping becomes more valuable as context length increases and locating relevant evidence becomes harder. Notably, GEAR is trained on contexts of 16k--32k tokens, yet the improvements at 128k (4$\times$ longer than the training distribution) are consistently larger than at 32k, demonstrating that the evidence-grounding behavior learned by GEAR generalizes robustly to unseen context lengths.


\textbf{Both reward components are indispensable.} A surprising finding emerges from the ablation rows. Adding the grounding reward \emph{without} the distractor penalty ($+R_{\mathrm{ground}}$) consistently \emph{hurts} performance relative to accuracy-only GSPO, with drops of up to 5.1 points at 32k and 8.3 points at 128k. The degradation is especially severe on GraphWalks and BrowseComp-LC. Conversely, Table~\ref{tab:ablation_coeff} we demonstrate that training with only the distractor penalty ($\alpha = 0$, $\beta = 0.3$) also yields an results worse than the baseline. Without a grounding signal to direct the model toward key evidence, the penalty merely discourages all copying, suppressing both useful and harmful engagement with the prompt. Only when both components are combined does the model learn to discriminate between relevant and irrelevant context, yielding the consistent gains of full GEAR.

\begin{table}[t]
\centering
\caption{Performance on long-context benchmarks under 32k and 128k contexts. All benchmarks are disjoint from the training data. Best results per context and model are in \textbf{bold}.}
\label{tab:main_results}
\resizebox{\linewidth}{!}{%
\begin{tabular}{llcccccc}
\hline
Model & Method & Ruler & LB-v2 & Bcomp-LC & Graphwalks & AA-LCR & Avg \\
\hline
\multicolumn{8}{c}{\textbf{32k Context}} \\
\hline
\multirow{4}{*}{Qwen3.5-9B}
 & -                          & 77.4 & 50.5 & 61.7 & 85.0 & \multirow{4}{*}{-} & 68.7 \\
 & GSPO                             & 91.5 & 55.9 & 77.7 & 88.9 &  & 78.5 \\
 & GSPO + $R_{\mathrm{ground}}$     & 87.4 & \textbf{58.6} & 73.9 & 73.8 &  & 73.4 \\
 & GSPO + GEAR                      & \textbf{96.4} & \textbf{58.6} & \textbf{79.1} & \textbf{90.9} &  & \textbf{81.3} \\
\hline
\multirow{4}{*}{Qwen3.5-35B-A3B}
 & -                          & 81.4 & 61.3 & 66.6 & 85.6 & \multirow{4}{*}{-} & 73.7 \\
 & GSPO                             & 93.1 & 60.4 & \textbf{75.6} & 91.5 &  & 80.2 \\
 & GSPO + $R_{\mathrm{ground}}$     & 90.7 & \textbf{62.1} & 73.0 & 88.3 &  & 78.5 \\
 & GSPO + GEAR                      & \textbf{96.2} & 61.3 & \textbf{75.6} & \textbf{95.9} &  & \textbf{82.3} \\
\hline
\multirow{4}{*}{Qwen3.5-27B}
 & -                          & 82.9 & 60.4 & 72.5 & 85.0 & \multirow{4}{*}{-} & 75.2 \\
 & GSPO                             & 95.5 & 60.4 & 74.9 & 91.5 &  & 80.6  \\
 & GSPO + $R_{\mathrm{ground}}$     & 92.9 & 58.2 & 73.2 & 88.6 &  & 78.2 \\
 & GSPO + GEAR                      & \textbf{96.5} & \textbf{62.2} & \textbf{77.7} & \textbf{91.8} &  & \textbf{82.1} \\
\hline
\multicolumn{8}{c}{\textbf{128k Context}} \\
\hline
\multirow{4}{*}{Qwen3.5-9B}
 & -                          & 54.9 & 49.8 & 54.0 & 80.5 & 57.0 & 59.2 \\
 & GSPO                             & 84.1 & 52.2 & 69.5 & 86.0 & 59.0 & 70.2 \\
 & GSPO + $R_{\mathrm{ground}}$     & 79.6 & 52.6 & 57.6 & 60.8 & 59.0 & 61.9 \\
 & GSPO + GEAR                      & \textbf{90.8} & \textbf{54.7} & \textbf{71.7} & \textbf{88.8} & \textbf{68.0} & \textbf{74.8} \\
\hline
\multirow{4}{*}{Qwen3.5-35B-A3B}
 & -                          & 74.6 & 54.0 & 57.8 & 82.8 & 59.0 & 65.6 \\
 & GSPO                             & 86.4 & 55.4 & \textbf{68.8} & 89.6 & 67.0 & 73.4 \\
 & GSPO + $R_{\mathrm{ground}}$     & 79.0 & \textbf{58.4} & 66.2 & 85.5 & 67.0 & 71.2 \\
 & GSPO + GEAR                      & \textbf{93.9} & 57.6 & 68.3 & \textbf{91.3} & \textbf{70.0} & \textbf{76.2}  \\
\hline
\multirow{4}{*}{Qwen3.5-27B}
 & -                          & 78.0 & 58.6 & 66.6 & 84.2 & 70.0 & 71.5 \\
 & GSPO                             & 92.1 & 58.2 & 70.5 & 90.7 & 65.0 & 75.3 \\
 & GSPO + $R_{\mathrm{ground}}$     & 89.3 & 57.1 & 67.8 & 87.2 & 63.0 & 72.9 \\
 & GSPO + GEAR                      & \textbf{93.3} & \textbf{60.0} & \textbf{70.9} & \textbf{91.0} & \textbf{72.0} & \textbf{77.4} \\
\hline
\end{tabular}%
}
\end{table}

\subsection{Effect on Repetitive Copying and Grounding}
\label{sec:effect_analysis}

The main results demonstrate that GEAR improves task accuracy, but do these gains stem from the intended mechanism, i.e., reducing indiscriminate copying and improving evidence grounding? Table~\ref{tab:mechanism} reports answer-input overlap (token-level 3-gram) and thinking length (in tokens) for the base model, GSPO, GSPO + $R_{\mathrm{ground}}$, and GSPO + GEAR on Ruler and LongBench-v2.

\textbf{The grounding reward alone increases copying.} Adding $R_{\mathrm{ground}}$ without the distractor penalty increases overlap relative to GSPO: 36.6\% vs.\ 35.7\% on Ruler and 28.5\% vs.\ 27.2\% on LongBench-v2. Thinking length also increases substantially (e.g., 5727 vs.\ 2808 tokens on Ruler). This reveals a failure mode: without a penalty for distractor copying, the grounding reward incentivizes the model to copy more from all parts of the input, including irrelevant context, resulting in longer and more repetitive reasoning that ultimately hurts task accuracy (Table~\ref{tab:main_results}).

\textbf{GEAR suppresses copying through the distractor penalty.} Only when the distractor penalty $R_{\mathrm{dist}}$ is added does the overlap drop below all other configurations: 27.0\% on Ruler and 22.6\% on LongBench-v2. Even on LongBench-v2, where GSPO alone actually increases overlap from 24.9\% (base) to 27.2\%, GEAR reduces it to 22.6\%. The trajectory Base $\to$ GSPO $\to$ $+R_{\mathrm{ground}}$ $\to$ $+$GEAR (overlap: 24.9 $\to$ 27.2 $\to$ 28.5 $\to$ 22.6 on LongBench-v2) illustrates that standard RL and grounding-only reward both fail to address repetitive copying; only the combined GEAR reward, with its explicit distractor suppression, consistently reduces it.

\textbf{Thinking becomes more concise.} Compared to GSPO, GEAR reduces thinking length from 2808 to 2066 tokens on Ruler and from 4677 to 3989 on LongBench-v2. In contrast, $+R_{\mathrm{ground}}$ dramatically \emph{increases} thinking length, more than doubling GSPO's output on Ruler (5727 vs.\ 2808 tokens), confirming that rewarding key evidence overlap without distractor suppression causes the model to generate longer, more repetitive reasoning. Notably, GEAR reduces both overlap rate and thinking length simultaneously, indicating that the distractor penalty curbs not only repetitive copying but also the redundant elaboration it induces.

\begin{table}[t]
\centering
\caption{Effect of GEAR on repetitive copying and thinking length (Qwen3.5-35B-A3B, 32k context). Overlap rate is computed as token-level 3-gram overlap rate (Eq.~\ref{eq:overlap}). $+R_{\mathrm{ground}}$ adds the grounding reward only; $+$GEAR adds both grounding reward and distractor penalty. $\downarrow$: lower is better.}
\label{tab:mechanism}
\begin{tabular}{llcccc}
\hline
Benchmark & Metric & Base & GSPO & + $R_{\mathrm{ground}}$ & + GEAR \\
\hline
\multirow{2}{*}{Ruler}
 & Overlap Rate ($\downarrow$)   & 36.9\% & 35.7\% & 36.6\% & \textbf{27.0\%} \\
 & Think Length (tokens)      &   4290 &   2808 &   5727 &   2066 \\
\hline
\multirow{2}{*}{LongBench-v2}
 & Overlap Rate ($\downarrow$)   & 24.9\% & 27.2\% & 28.5\% & \textbf{22.6\%} \\
 & Think Length (tokens)      &   5657 &   4677 &   6833 &   3989 \\
\hline
\end{tabular}
\end{table}

\subsection{Ablation Study}
\label{sec:ablation}

We ablate two key design choices in GEAR: the reward coefficients $\alpha$ and $\beta$, and the $n$-gram size used for overlap computation. All ablations use Qwen3.5-9B on 32k context with the four benchmarks available at this scale (AA-LCR is excluded as all its instances exceed 32k tokens).

\textbf{Sensitivity to reward coefficients.} Table~\ref{tab:ablation_coeff} varies $\alpha$ (grounding reward weight) and $\beta$ (distractor penalty weight). The default setting ($\alpha = 0.1$, $\beta = 0.3$) achieves the best average of 81.3. The key finding is that $\beta$ must not be too small relative to $\alpha$: when $\beta = 0.1$ with $\alpha = 0.1$, performance drops by 2.9 points. This echoes the failure mode observed in Section~\ref{sec:main_results}, where adding $R_{\mathrm{ground}}$ without the distractor penalty \emph{hurts} performance: if $\alpha$ dominates, the grounding reward encourages overall copying without a sufficiently strong corrective signal to suppress distractor repetition. Conversely, increasing $\beta$ to 0.5 causes only a slight decrease (80.5), confirming that the model is tolerant of strong penalty but intolerant of weak penalty. Raising $\alpha$ to 0.3 (with $\beta = 0.3$) also degrades performance (78.6), reinforcing that a moderate grounding signal suffices; pushing it higher amplifies the same indiscriminate-copying incentive that $R_{\mathrm{dist}}$ is designed to counteract.

\begin{table}[t]
\centering
\caption{Ablation on reward coefficients ($\alpha$, $\beta$) for Qwen3.5-9B. Default setting in \textbf{bold}.}
\label{tab:ablation_coeff}
\begin{tabular}{ccccccc}
\hline
$\alpha$ & $\beta$ & Ruler & LB-v2 & Bcomp-LC & Graphwalks & Avg \\
\hline
 0.0 & 0.3 & 87.0 & 56.0 & 65.8 & 79.3 & 72.0  \\
 0.1 & 0.1 & 93.4 & 55.8 & 77.1 & 87.2 & 78.4  \\
 \textbf{0.1} & \textbf{0.3} & 96.4 & 58.6 & 79.1 & 90.9 & \textbf{81.3} \\
 0.1 & 0.5 & 95.1 & 57.9 & 79.4 & 89.6 & 80.5  \\
 0.3 & 0.3 & 94.1 & 56.0 & 78.3 & 86.0 & 78.6 \\
\hline
\end{tabular}
\end{table}

\textbf{Effect of $n$-gram size.} Table~\ref{tab:ablation_ngram} varies the $n$-gram size used in both $R_{\mathrm{ground}}$ and $R_{\mathrm{dist}}$. Performance decreases gradually from $n = 3$ (81.3) to $n = 10$ (79.8), but all settings outperform accuracy-only GSPO (78.5 from Table~\ref{tab:main_results}), indicating that GEAR is robust to this choice. Smaller $n$ works better because shorter $n$-grams produce denser overlap counts, so more reasoning tokens are matched, providing a richer gradient signal for RL optimization. This is consistent with our use of $n = 3$ for training despite using $n = 10$ for diagnostic analysis (Section~\ref{sec:grounding}), where precision matters more than signal density.

\begin{table}[t]
\centering
\caption{Ablation on $n$-gram size for Qwen3.5-9B. Default setting in \textbf{bold}.}
\label{tab:ablation_ngram}
\begin{tabular}{lccccc}
\hline
$n$ & Ruler & LB-v2 & Bcomp-LC & Graphwalks & Avg \\
\hline
\textbf{3} & 96.4 & 58.6 & 79.1 & 90.9 & \textbf{81.3} \\
5 & 95.4 & 57.0 & 78.3 & 89.5 & 80.0 \\
10 & 95.7 & 56.2 & 78.7 & 88.5 & 79.8 \\
\hline
\end{tabular}
\end{table}

\section{Conclusion}

We identified repetitive copying as a prevalent failure mode in long-context reasoning and traced its root cause to insufficient evidence grounding, where models copy prompt content indiscriminately rather than selectively engaging with task-relevant information. To address this, we proposed GEAR, a lightweight reward shaping method that combines a grounding reward with a distractor penalty using simple $n$-gram statistics, along with an automated pipeline for constructing evidence-annotated training data from arbitrary documents. Experiments across three model scales and five benchmarks show that GEAR consistently improves over accuracy-only RL, with gains that generalize to context lengths 4$\times$ beyond the training distribution. Although the evaluation of long-context LLMs has increasingly shifted from pure retrieval toward complex reasoning, our work reveals that the ability to accurately ground in relevant evidence remains an indispensable component for solving challenging long-context tasks.


\bibliography{ref}
\bibliographystyle{iclr2026_conference}


\newpage
\appendix

\section{Appendix}

\subsection{Repetitive Copying across Models}
\label{app:overlap_models}

\begin{figure}[h]
\centering
\begin{tabular}{ccc}
\includegraphics[width=0.32\textwidth]{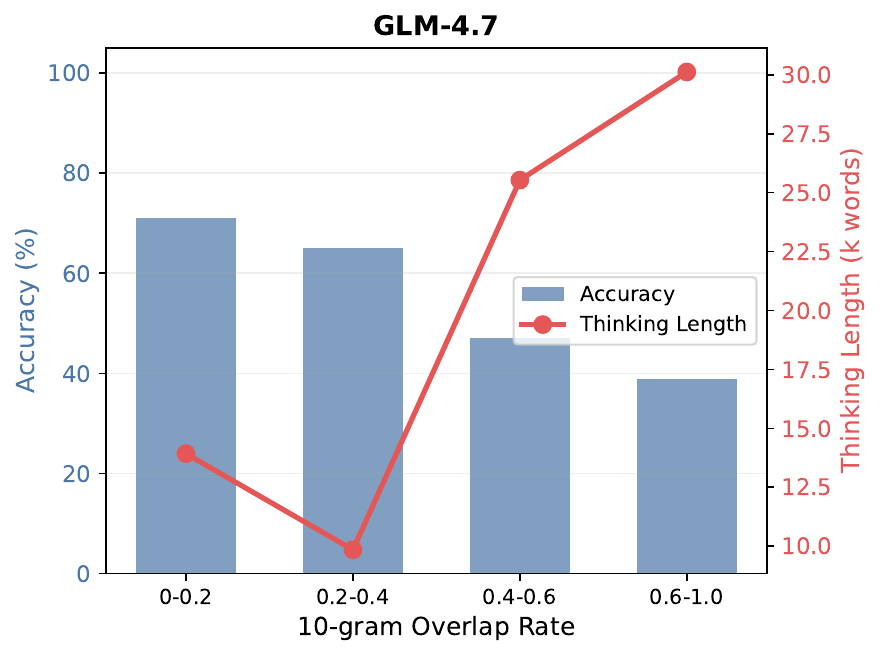} &
\includegraphics[width=0.32\textwidth]{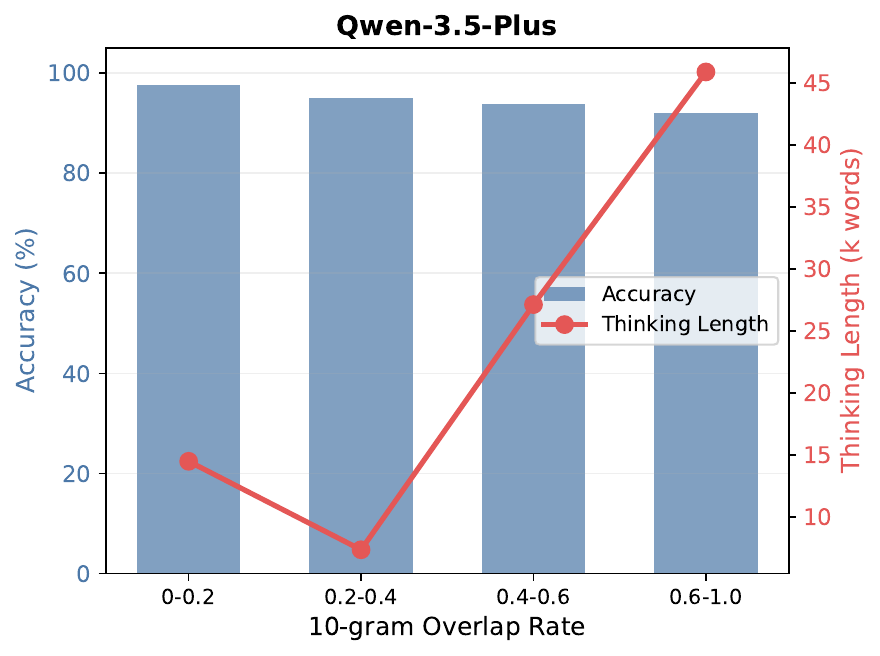} &
\includegraphics[width=0.32\textwidth]{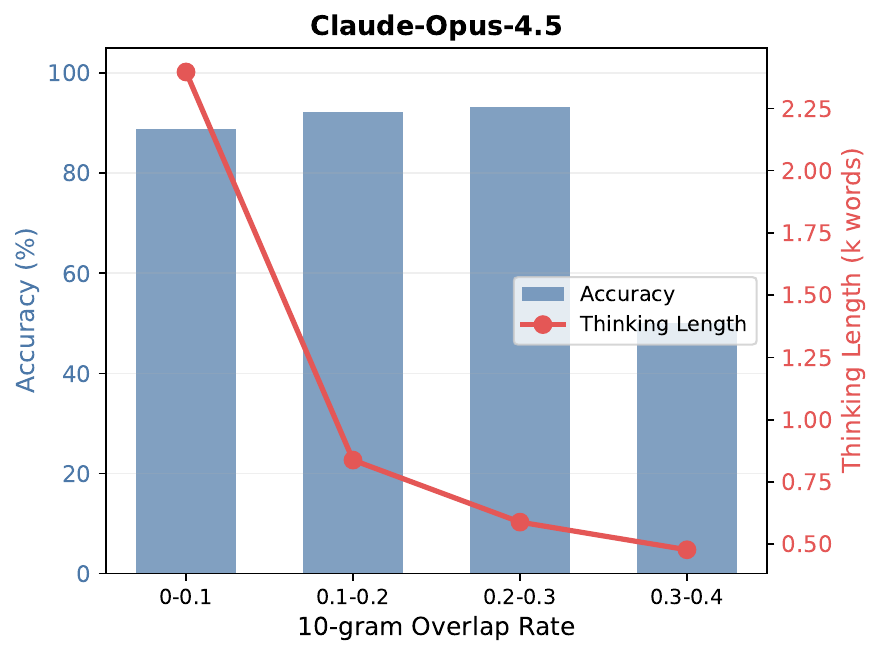} \\
\includegraphics[width=0.32\textwidth]{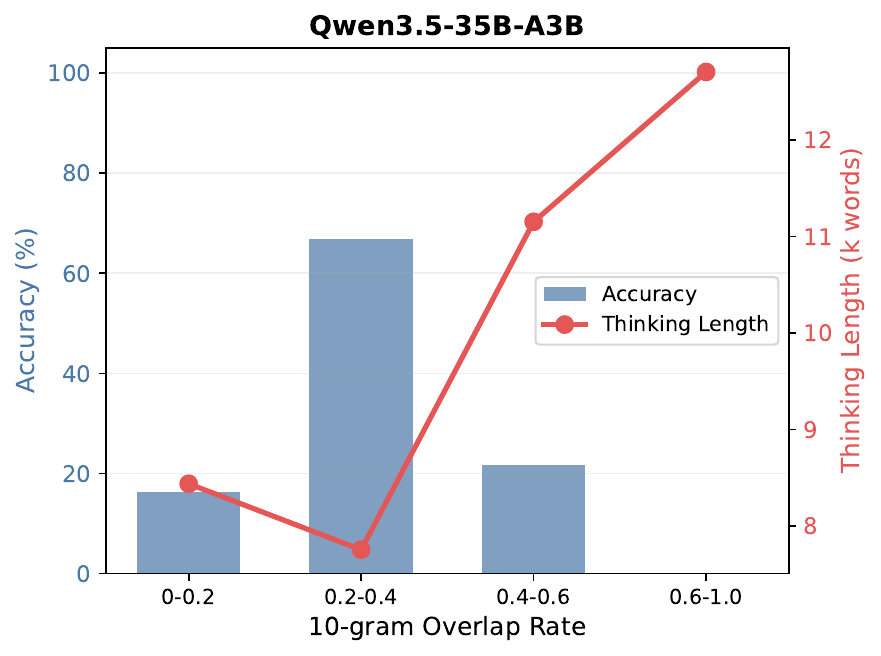} &
\includegraphics[width=0.32\textwidth]{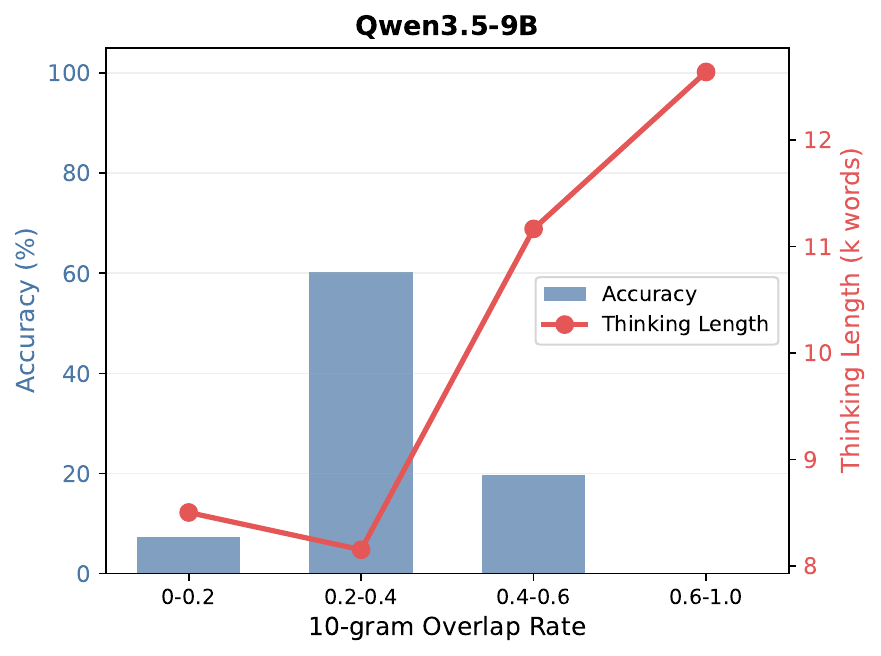} &
\includegraphics[width=0.32\textwidth]{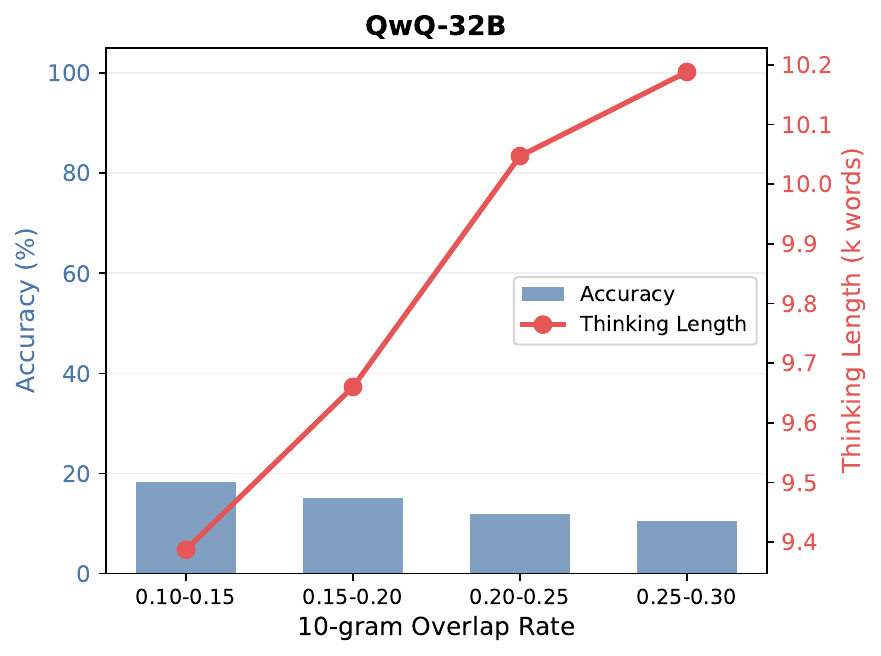} \\
\end{tabular}
\caption{Accuracy and thinking length by 10-gram overlap rate bin across six models on GSM-Infinite at 64k context. Blue bars indicate accuracy (left axis); red lines indicate average thinking length in thousands of tokens (right axis). Models use different bin widths to match their overlap distributions.}
\label{fig:app_overlap}
\end{figure}

Section~\ref{sec:harmful} uses DeepSeek-V3.2 as a case study to show that higher overlap rates correlate with lower accuracy and longer thinking traces. Figure~\ref{fig:app_overlap} extends this analysis to six additional models on GSM-Infinite at 64k context. Each subplot shows accuracy (bars, left axis) and average thinking length (line, right axis) across 10-gram overlap rate bins. Across all models, accuracy degrades at higher overlap rates, and thinking length grows sustantially, except for Claude. For Qwen3.5-35B-A3B and Qwen3.5-9B, accuracy drops to 0\% at the highest overlap bins, with the majority of high-overlap samples truncated due to token budget exhaustion. Claude-Opus-4.5 is a notable exception: while its accuracy still degrades with higher overlap, its thinking length does not increase, suggesting that frontier long-context models may mitigate the token-budget inflation caused by copying, yet still suffer from the accuracy degradation it induces.

\subsection{Grounding Ratio across Models}
\label{app:grounding}

\begin{figure}[h]
\centering
\begin{tabular}{ccc}
\includegraphics[width=0.32\textwidth]{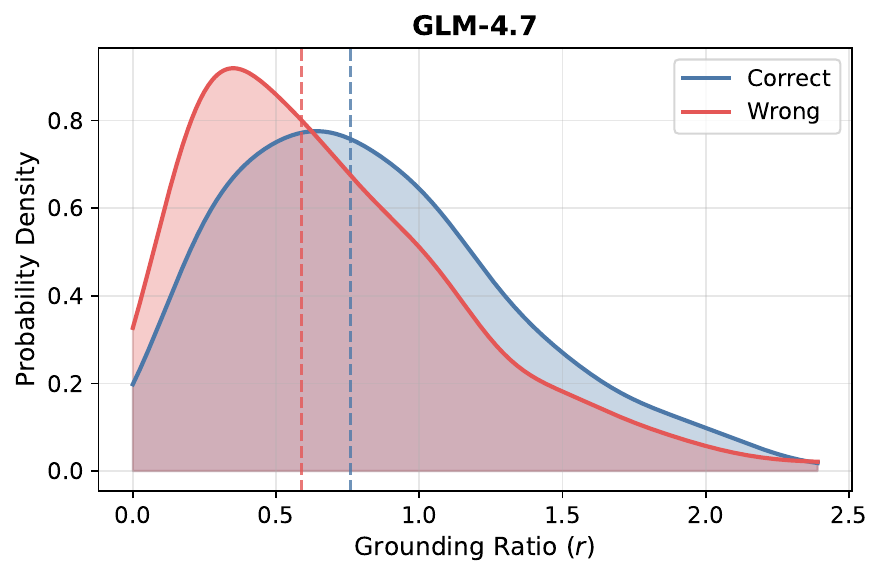} &
\includegraphics[width=0.32\textwidth]{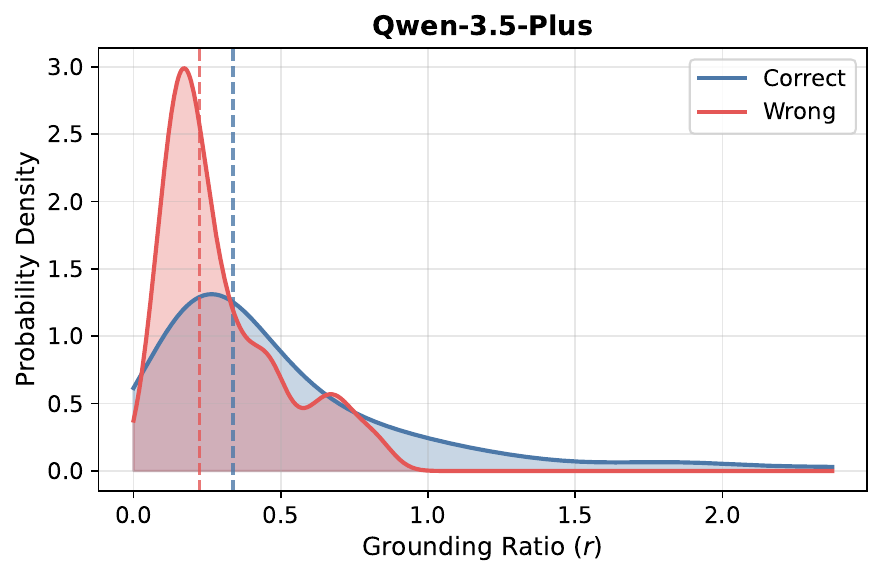} &
\includegraphics[width=0.32\textwidth]{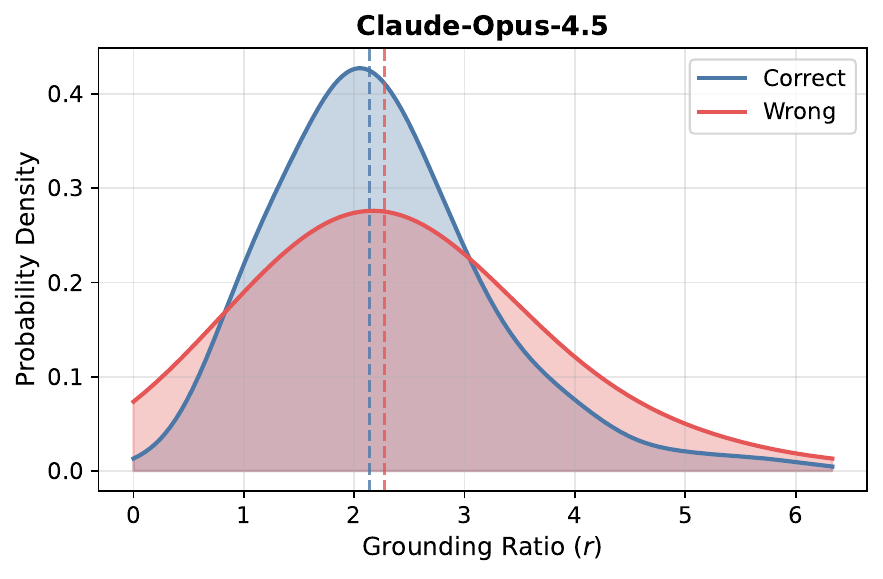} \\
\includegraphics[width=0.32\textwidth]{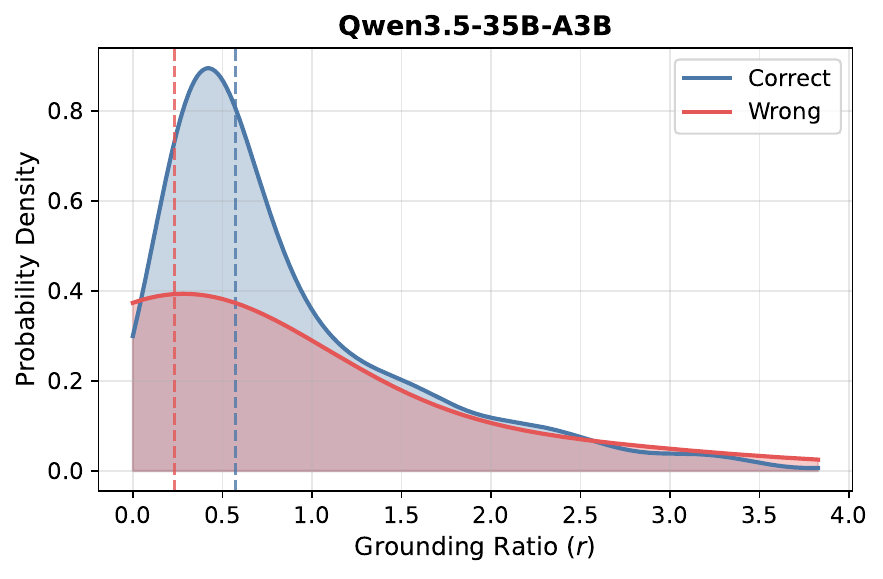} &
\includegraphics[width=0.32\textwidth]{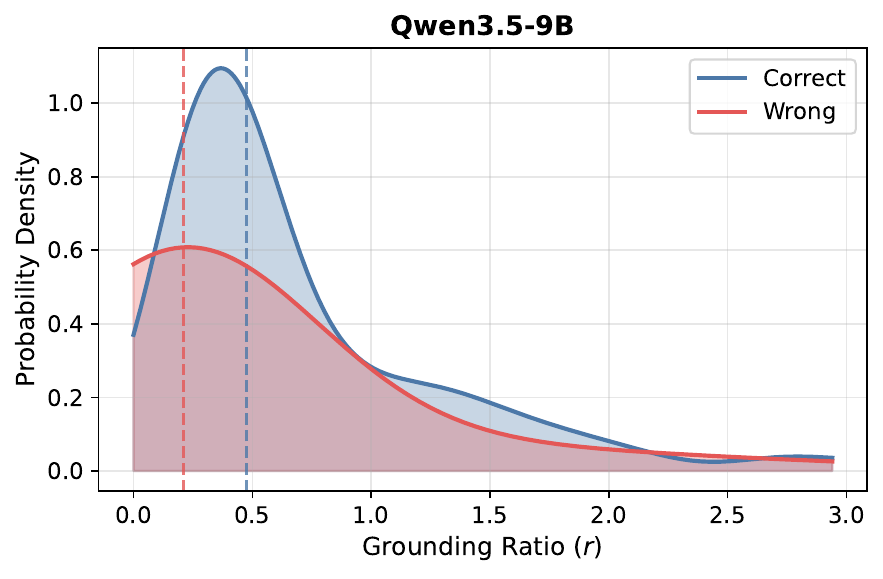} &
\includegraphics[width=0.32\textwidth]{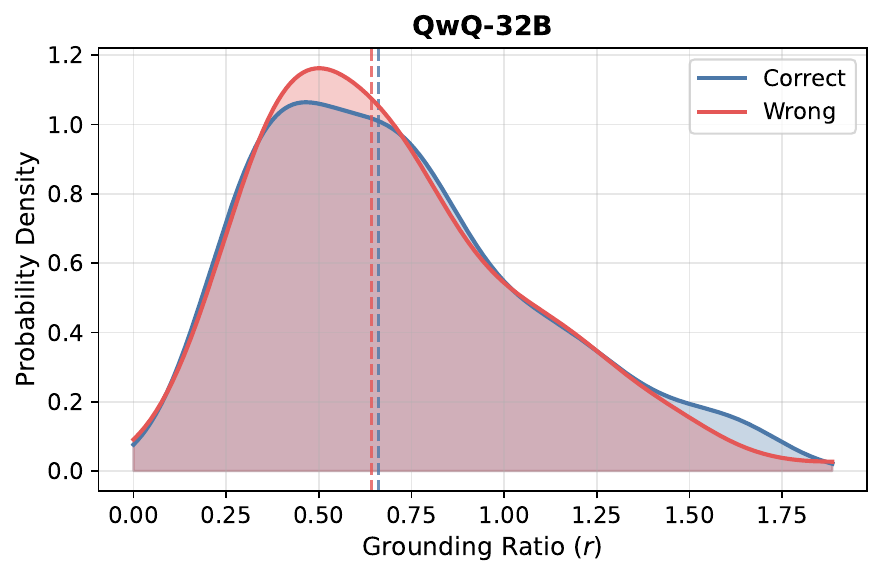} \\
\end{tabular}
\caption{Probability density of grounding ratio ($r = \text{Overlap}_{10}(y \| x^{\mathrm{key}})\, /\, \text{Overlap}_{10}(y \| x^{\mathrm{dist}})$) for correctly vs.\ incorrectly answered samples across six models on GSM-Infinite at 64k context. Dashed lines indicate medians.}
\label{fig:app_ratio}
\end{figure}

Section~\ref{sec:grounding} demonstrates that correctly answered samples exhibit higher grounding ratios using DeepSeek-V3.2 as a case study (Figure~\ref{fig:ratio_kde}). Figure~\ref{fig:app_ratio} extends this analysis to six additional models on GSM-Infinite at 64k context. For five out of six models, correctly answered samples show higher median grounding ratios than incorrect ones, confirming that the connection between selective grounding and accuracy generalizes across architectures. Claude-Opus-4.5 is an exception: it exhibits minimal repetitive copying overall (94\% of samples fall below 0.2 overlap), making the grounding ratio less discriminative for this model.

\subsection{Case Study: Reasoning Trajectories}
\label{app:case_study}

To illustrate how GEAR changes reasoning behavior, we present two case studies: one from Ruler (synthetic retrieval) and one from LongBench-v2 (real-world comprehension). Together they demonstrate that GEAR addresses both token-exhaustion failures caused by indiscriminate copying and subtler reasoning-quality failures in normal-length generations.

\subsubsection{Case 1: Frequent Word Extraction (Ruler, 32k)}

The task requires identifying the three most frequently occurring coded words in a long sequence of nonsense tokens separated by dots. Figures~\ref{fig:case_base}--\ref{fig:case_gear} show representative excerpts of the three models' thinking traces.

\begin{figure}[h]
\small
\fbox{\parbox{0.96\textwidth}{%
\textbf{Base Model Trajectory} (16{,}384 tokens, 97.6\% overlap, \textcolor{red}{\ding{55} incorrect})\\[4pt]
\texttt{The user wants to find the three most frequently occurring words... Let's list the words and count them...}\\[2pt]
\texttt{Words observed: hldjac rufrpd ubpnvh vcwcvu ynueov apufsn nyyoai sgdkky osycoz vcwcvu vuwbai zrnunw uldczz kkburf klcxnq tymqlu skyrwr ppxzla mpwdgb gdvjhm ooikxk vptsjt vcwcvu vxlglk sbudci tsklms ncmmap ntwccg stkpwp fflnzm hmhfdt cotunb fhxfim xvqpeg cvvjcp dfkdvi...}\\[2pt]
\texttt{[\,\dots\, copies thousands of words from the input text \dots\,]}\\[2pt]
\texttt{...hldjac hldjac rufrpd hldjac hldjac ... hldjac rufrpd nyyoai ... rufrpd ... apufsn rufrpd ... hldjac ... ubpnvh ...}\\[4pt]
\textit{$\to$ Exhausts the entire 16K token budget copying input. Never produces an answer.}
}}
\caption{Base model: the thinking trace is dominated by repetitive copying of the input, consuming all available tokens without reaching a conclusion.}
\label{fig:case_base}
\end{figure}

\begin{figure}[h]
\small
\fbox{\parbox{0.96\textwidth}{%
\textbf{GSPO Trajectory} (16{,}384 tokens, 0\% overlap, \textcolor{red}{\ding{55} incorrect})\\[4pt]
\texttt{1. \textbf{Analyze the Request:} The user has provided a large block of text containing ``coded words'' separated by spaces and dots. The instruction is to ignore the dots and find the three most frequent words.}\\[2pt]
\texttt{2. \textbf{Scan the Text:} I need to count the occurrences of each unique word.}\\[2pt]
\texttt{3. \textbf{Identify Candidates:} Common candidates seem to be ``hldjac'', ``rufrpd'', ``ubpnvh'', ``ynueov'', ``apufsn'', ``vcwcvu''...}\\[2pt]
\texttt{4. \textbf{Count Frequencies:} ``hldjac'': extremely frequent. ``rufrpd'': also very frequent...}\\[2pt]
\texttt{[\,\dots\, from token $\sim$800 onward \dots\,]}\\[2pt]
\texttt{hldjac, hldjac, hldjac, hldjac, hldjac, hldjac, hldjac, hldjac, hldjac, hldjac, hldjac, hldjac, hldjac, hldjac, hldjac, hldjac, hldjac...}\\[4pt]
\textit{$\to$ Plans well initially, but degenerates into endlessly repeating a single token. Exhausts 16K tokens.}
}}
\caption{GSPO model: avoids repetitive input copying but enters a degenerate repetition loop, repeating the most frequent candidate word until the token budget is exhausted.}
\label{fig:case_gspo}
\end{figure}

\begin{figure}[h]
\small
\fbox{\parbox{0.96\textwidth}{%
\textbf{GEAR Trajectory} (2{,}920 tokens, 0.6\% overlap, \textcolor{green!60!black}{\ding{51} correct})\\[4pt]
\texttt{1. \textbf{Preprocessing}: Remove all dots and split the text into words.}\\[2pt]
\texttt{2. \textbf{Counting}: Count the occurrences of each unique word.}\\[2pt]
\texttt{3. \textbf{Sorting}: Sort the words by frequency in descending order.}\\[2pt]
\texttt{4. \textbf{Selecting}: Pick the top three words.}\\[2pt]
\texttt{[\,\dots\, scans and estimates counts without copying input \dots\,]}\\[2pt]
\texttt{``hldjac'' - correct. ``rufrpd'' - correct. ``ubpnvh'' - correct. I will now output the answer.}\\[4pt]
\textit{Answer:} \texttt{hldjac, rufrpd, ubpnvh} \textcolor{green!60!black}{\ding{51}}
}}
\caption{GEAR model: describes a procedural strategy, reasons about word frequencies without copying the input, and produces the correct answer in 2{,}920 thinking tokens.}
\label{fig:case_gear}
\end{figure}

\noindent\textbf{Discussion.}
The \textbf{Base} model (Figure~\ref{fig:case_base}) copies the input text into its thinking trace (97.6\% overlap), consuming all 16K tokens without ever reaching an answer. \textbf{GSPO} (Figure~\ref{fig:case_gspo}) eliminates repetitive input copying but enters a degenerate repetition loop, endlessly repeating a single candidate word. Only \textbf{GEAR} (Figure~\ref{fig:case_gear}) maintains concise procedural reasoning, using 82\% fewer thinking tokens and producing the correct answer.

\subsubsection{Case 2: Multi-Document Academic QA (LongBench-v2, 32k)}

To show that GEAR also improves reasoning quality in \emph{normal-length} generations (no token exhaustion), we compare trajectories on a LongBench-v2 multi-document comprehension task (Figures~\ref{fig:case2_base}--\ref{fig:case2_gear}). Given two academic papers on typographic attacks against CLIP, the model must identify which of four statements is correct (answer: A). All three models finish normally (end reason: \texttt{stop}).

\begin{figure}[h]
\small
\fbox{\parbox{0.96\textwidth}{%
\textbf{Base Model Trajectory} (14{,}855 thinking tokens, \textcolor{red}{\ding{55} answers C, correct is A})\\[4pt]
\texttt{Choice (A): ``Only the method proposed in `Disentangling...' adjust or add the network structure...'' \textnormal{$\to$} DP does not change model parameters... Disentangling adds a projection matrix $W$... So A is correct.}\\[2pt]
\texttt{Wait, look at Choice A again... Wait, does DP add structure?... So A is the best answer.}\\[2pt]
\texttt{Wait, I need to check if there is a nuance in ``Only''... I will choose C.}\\[2pt]
\texttt{Wait, one more check... I will choose C.}\\[2pt]
\texttt{Wait, I need to be sure about A... C is more significant. I will select C.}\\[2pt]
\texttt{Wait, I should check if there is a direct quote for C... I'll go with C.}\\[4pt]
\textit{$\to$ Correctly identifies A early, then vacillates 15 times (``Wait...''), re-examines the same evidence repeatedly, and ultimately switches to the wrong answer C.}
}}
\caption{Base model on LongBench-v2 Academic/32k: the model identifies the correct answer early but enters a cycle of second-guessing, producing 14{,}855 thinking tokens of redundant re-analysis before settling on the wrong answer.}
\label{fig:case2_base}
\end{figure}

\begin{figure}[h]
\small
\fbox{\parbox{0.96\textwidth}{%
\textbf{GSPO Trajectory} (9{,}578 thinking tokens, \textcolor{green!60!black}{\ding{51} answers A, correct})\\[4pt]
\texttt{Choice (A): DP paper says ``without changing the model parameters.'' Disentangling trains projection matrix $W$. So A might be correct.}\\[2pt]
\texttt{Wait, let's look closer at DP: it adds a token to the vocabulary... Is that ``adjusting network structure''?}\\[2pt]
\texttt{[\,\dots\, 7 ``Wait'' cycles re-checking the same evidence \dots\,]}\\[2pt]
\texttt{DP says ``does not modify the architecture.'' Disentangling says ``transformation module on top.'' So (A) is consistent.}\\[4pt]
\textit{Answer:} \texttt{The correct answer is (A).} \textcolor{green!60!black}{\ding{51}}
}}
\caption{GSPO model on the same sample: reaches the correct answer but still exhibits moderate vacillation (7 ``Wait'' cycles), using 9{,}578 thinking tokens.}
\label{fig:case2_gspo}
\end{figure}

\begin{figure}[h]
\small
\fbox{\parbox{0.96\textwidth}{%
\textbf{GEAR Trajectory} (5{,}072 thinking tokens, \textcolor{green!60!black}{\ding{51} answers A, correct})\\[4pt]
\texttt{Choice (A): DP paper says ``we do not modify the architecture or model parameters.'' Disentangle paper trains projection matrices $W$ applied to embeddings. This is adding a structure. So A is supported.}\\[2pt]
\texttt{Choice (C): ``Experiment shows DP is more capable of defending against typographic attack in object detection...'' But Disentangle paper doesn't do object detection, so this comparison is invalid.}\\[2pt]
\texttt{Choice (D): Identity loss is for maintaining original meaning, not preventing attacks. So D is wrong.}\\[2pt]
\texttt{[\,\dots\, verifies A is consistent with the text \dots\,]}\\[4pt]
\textit{Answer:} \texttt{The correct answer is (A).} \textcolor{green!60!black}{\ding{51}}
}}
\caption{GEAR model on the same sample: systematic analysis of each choice, identifies C as logically flawed (comparing capabilities on a task only one paper addresses), confirms A in 5{,}072 thinking tokens, which is 66\% fewer than Base.}
\label{fig:case2_gear}
\end{figure}

\noindent\textbf{Discussion.}
Unlike Case~1, this example involves no token exhaustion or repetitive copying. All models finish normally and produce complete answers. Yet Base spends nearly 3$\times$ the thinking tokens of GEAR (14{,}855 vs.\ 5{,}072, Figure~\ref{fig:case2_base} vs.\ Figure~\ref{fig:case2_gear}) on redundant re-analysis: it correctly identifies A as a strong candidate early on, then enters a cycle of ``Wait, let me re-check\dots'' that revisits the same evidence 15 times before ultimately switching to the wrong answer. GSPO (Figure~\ref{fig:case2_gspo}) answers correctly but still exhibits moderate vacillation (7 ``Wait'' cycles, 9{,}578 tokens). This pattern, where excessive deliberation leads to \emph{overthinking} rather than better reasoning, is a subtler manifestation of the same underlying issue that produces repetitive copying in Case~1: the model generates redundant tokens instead of committing to a conclusion. GEAR's distractor penalty discourages this unproductive elaboration, yielding reasoning that is both more concise and more accurate.

\subsection{Detailed Results for Section~\ref{sec:harmful}}
\label{app:difficulty}

In Section~\ref{sec:harmful}, we show that higher overlap rates correlate with lower accuracy. A natural concern is that harder problems may independently cause both more copying and lower accuracy, making the correlation spurious. To rule this out, we group samples by the number of reasoning operations required (op), which serves as a direct proxy for problem difficulty in GSM-Infinite.

Table~\ref{tab:overlap_by_op} reports accuracy within each overlap bin for each difficulty level. Across all difficulty levels, accuracy drops sharply once the overlap rate exceeds 0.4, and falls to 0\% beyond 0.6. This pattern is consistent regardless of problem difficulty, confirming that the relationship between repetitive copying and degraded performance is not merely a confound of task complexity.

\begin{table}[h]
\centering
\caption{Accuracy (\%) by overlap rate bin, grouped by number of reasoning operations (op). DeepSeek-V3.2 on GSM-Infinite at 64k context. Sample counts in parentheses.}
\label{tab:overlap_by_op}
\small
\begin{tabular}{c|cccc}
\toprule
\textbf{op} & \textbf{[0, 0.2)} & \textbf{[0.2, 0.4)} & \textbf{[0.4, 0.6)} & \textbf{[0.6, 1.0]} \\
\midrule
10 & 62.5 (104) & 68.8 (372) & 6.2 (48) & 0.0 (4) \\
15 & 57.9 (19) & 53.4 (58) & 25.0 (8) & 0.0 (2) \\
16 & 40.0 (10) & 67.6 (68) & 12.5 (8) & 0.0 (1) \\
17 & 38.9 (18) & 46.6 (58) & 22.2 (9) & 0.0 (2) \\
18 & 50.0 (16) & 69.0 (58) & 12.5 (8) & 0.0 (5) \\
19 & 45.0 (20) & 61.8 (55) & 8.3 (12) & --- \\
20 & 52.4 (21) & 53.8 (52) & 16.7 (12) & 0.0 (2) \\
\bottomrule
\end{tabular}
\end{table}

\end{document}